\documentclass[letterpaper, conference]{IEEEtran}
\IEEEoverridecommandlockouts
\usepackage{cite}
\usepackage{amsmath,amssymb,amsfonts}
\usepackage{algorithmic}
\usepackage{graphicx}
\usepackage{textcomp}
\usepackage{xcolor}
\usepackage{booktabs}
\usepackage{makecell}
\usepackage{subfigure}
\usepackage{multirow}
\usepackage{url}
\usepackage{dblfloatfix}
\usepackage{gensymb}
\usepackage{makecell}
\usepackage{comment}
\usepackage{environ} 

\usepackage{enumerate}

\newcolumntype{C}[1]{>{\centering\let\newline\\\arraybackslash\hspace{0pt}}m{#1}}
\NewEnviron{NORMAL}{%
    \scalebox{0.9}{$\BODY$} 
} 
\usepackage[ruled,vlined]{algorithm2e}

\usepackage{enumerate}

\def\BibTeX{{\rm B\kern-.05em{\sc i\kern-.025em b}\kern-.08em
    T\kern-.1667em\lower.7ex\hbox{E}\kern-.125emX}}
\begin{document}
\bstctlcite{IEEEexample:BSTcontrol}
\title{Autonomous Navigation and Configuration of Integrated Access Backhauling for UAV Base Station Using Reinforcement Learning}

\author{
    \begin{tabular}{c}\IEEEauthorblockN{Hongyi Zhang\IEEEauthorrefmark{1}, Jingya Li\IEEEauthorrefmark{2}, Zhiqiang Qi\IEEEauthorrefmark{2}, Xingqin Lin\IEEEauthorrefmark{2}, Anders Aronsson\IEEEauthorrefmark{2},\\ Jan Bosch\IEEEauthorrefmark{1}, Helena Holmström Olsson\IEEEauthorrefmark{3} \end{tabular}\\}
    \IEEEauthorblockA{\IEEEauthorrefmark{1}\textit{Chalmers University of Technology}, Gothenburg, Sweden. Email: \{hongyiz, jan.bosch\}@chalmers.se} 
    
    \IEEEauthorblockA{\IEEEauthorrefmark{2}\textit{Ericsson Research}, Ericsson. Email: \{jingya.li, zhiqiang.qi, xingqin.lin, anders.aronsson\}@ericsson.com}

    \IEEEauthorblockA{\IEEEauthorrefmark{3}\textit{Malmö University}, Malmö, Sweden. Email: helena.holmstrom.olsson@mau.se}
    
    \vspace{-20pt}

}

\maketitle

\begin{abstract}

Fast and reliable connectivity is essential to enhancing situational awareness and operational efficiency for public safety mission-critical (MC) users. In emergency or disaster circumstances, where existing cellular network coverage and capacity may not be available to meet MC communication demands, deployable-network-based solutions such as cells-on-wheels/wings can be utilized swiftly to ensure reliable connection for MC users. In this paper, we consider a scenario where a macro base station (BS) is destroyed due to a natural disaster and an unmanned aerial vehicle carrying BS (UAV-BS) is set up to provide temporary coverage for users in the disaster area. The UAV-BS is integrated into the mobile network using the 5G integrated access and backhaul (IAB) technology. We propose a framework and signalling procedure for applying machine learning to this use case. A deep reinforcement learning algorithm is designed to jointly optimize the access and backhaul antenna tilt as well as the three-dimensional location of the UAV-BS in order to best serve the on-ground MC users while maintaining a good backhaul connection. Our result shows that the proposed algorithm can autonomously navigate and configure the UAV-BS to improve the throughput and reduce the drop rate of MC users.

\end{abstract}


\vspace{-5pt}
\section{Introduction}

Like food, water and medicine, the ability to communicate has proven to be an essential tool for first responders, governments, and survivors in disaster response and relief. To provide connectivity in areas that cannot be fully covered by the existing mobile network, for example, when the network infrastructure is damaged or not available, unmanned aerial vehicles (UAVs) carrying base stations (BSs) can be used to provide temporary coverage for users located in the disaster area. 

UAV-BS assist wireless communication networks have recently gained increased interest in both academic and public safety communities\cite{li20215g, naqvi2018drone, firstnet, merwaday2016improved, ferranti2020skycell}. Thanks to the great mobility and flexibility of UAVs, it is expected that UAV-BSs can bring fast connectivity for mission-critical (MC) communications. However, there are a number of challenges that must be addressed when deploying UAV-BSs in practice. The deployment and configuration of the UAV-BSs play a critical role in the performance of the target services. When integrating a UAV-BS into an existing mobile network, a fast and reliable backhaul connection between the UAV-BS and on-ground BSs is required to ensure the end-to-end quality of service (QoS) for the interested users. In addition, reliable and scalable backhaul links between different UAV-BSs are needed when multiple UAV-BSs are used to cover a wider area. Therefore, it is crucial to ensure the good quality of both the access and backhaul links when optimizing the deployment of UAV-BSs. The deployment optimization also depends on many other factors such as the limitations on UAV’s flying altitude, operation time, antenna capabilities and transmit power, the network traffic load distribution, and user movements. 

While many works on UAV-BS deployment focused on the problems of UAV placement, trajectory design, and number of UAV-BSs, etc., only a few previous results have considered the wireless backhaul aspects\cite{wang2019deployment,kalantari2017backhaul,cicek2020backhaul,tafintsev2020aerial}.
In \cite{wang2019deployment}, the authors investigated how to rapidly deploy the minimum number of UAV-BSs to assist the existing mobile network to evenly serve as many users as possible while guaranteeing a robust wireless connection among the UAV-BSs and fixed on-ground BSs. It is assumed that all UAV-BSs are flying at the same and fixed height, and the robustness of the backbone network among the deployed UAV-BSs is guaranteed by ensuring a bi-connection network topology so that if one UAV-BS fails, there still exists at least one route between any UAV-BS and a fixed on-ground BS. In \cite{kalantari2017backhaul}, a UAV-BS 3-D placement algorithm is proposed to maximize the total number of served users or the sum of user data rates subject to capacity constraints of both access and backhaul links. This work was further extended by the authors in \cite{cicek2020backhaul}, where a mixed-integer non-linear programming approach is proposed to jointly optimize a UAV-BS location and the system bandwidth allocation without exceeding the backhaul and access capacities. 

\begin{figure*}[t]
  \begin{center}
    \includegraphics[scale=0.52]{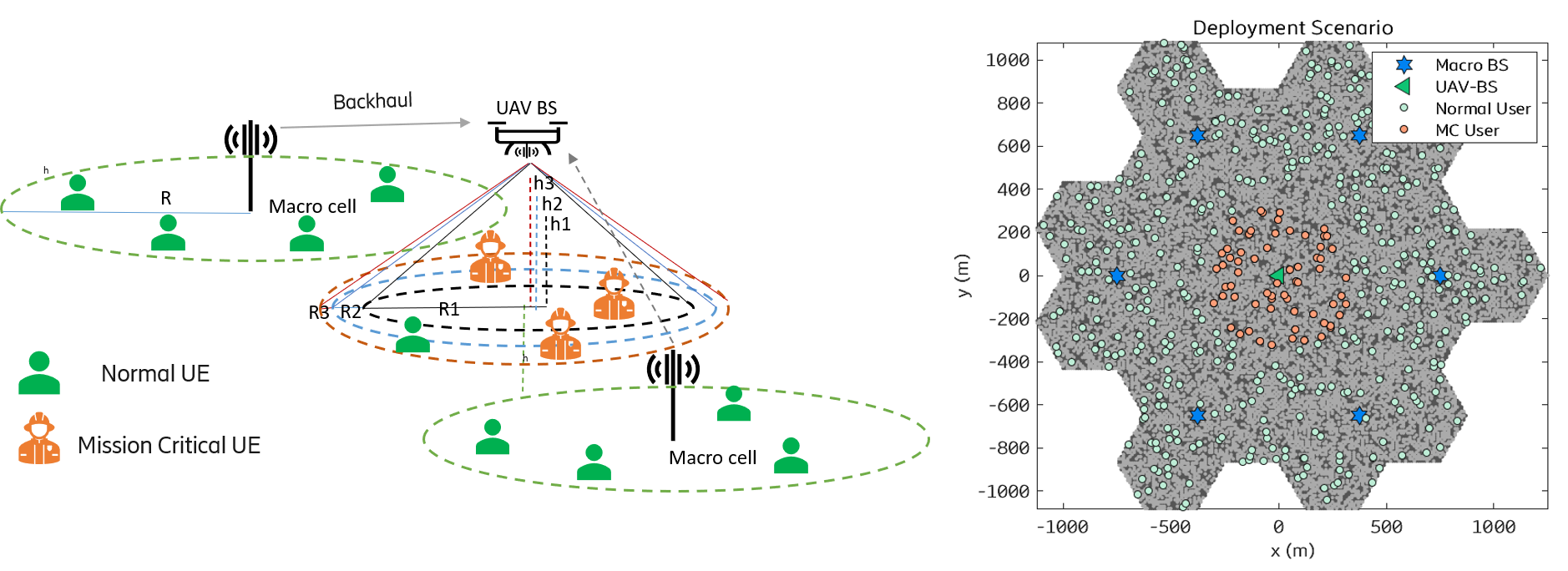}
    \setlength{\abovecaptionskip}{-5pt}
    \caption{A UAV-BS assisted wireless network deployment using IAB}
    \vspace{-15pt}
    \label{fig:DeploymentScenario}
  \end{center}
\end{figure*}

With 5G new radio (NR), there is an opportunity to use the integrated access and backhaul (IAB) feature to wirelessly connect multiple UAV-BSs and integrate them to an existing mobile network seamlessly \cite{li20215g,tafintsev2020aerial}. The NR IAB feature supports multi-hop wireless backhaul with a flexible and adaptive network architecture \cite{madapatha2020integrated}. Figure 1 illustrates an example of UAV-BS-assisted network deployment using IAB. A macro-BS with a wired connection to the core network is configured as an IAB donor node, and a UAV-BS is configured as an IAB node. The UAV-BS connects to a parent or donor node using wireless backhaul, and it services on-ground users using access links. The IAB network topology can adapt to the varying backhaul link conditions and traffic load situations.

In \cite{tafintsev2020aerial}, the authors evaluated the mean user throughput and user fairness performance of a UAV-based IAB system in millimeter-wave (mmWave) urban deployments, where the UAV-BS 2-D location is optimized to follow the user movement using a particle swarm optimization method. They assumed separate channels for access and backhaul links as well as dedicated antenna arrays for each interface. In this work, we consider a UAV-BS-assisted IAB network for providing temporary coverage to MC users in an emergency area. We assume that the system operates in a mid-band, which provides better coverage than mmWave bands. The same frequency band and the same antennas are shared between access and backhaul links to reduce the cost and weight of the BS carried on the UAV. We investigate how reinforcement learning (RL) can support autonomous navigation and configuration of IAB for UAV-BS-assisted networks. A framework and signalling procedure are proposed to support applying RL in an IAB network architecture. In addition, an RL algorithm is designed to jointly optimize the antenna configuration and the 3-D location of the UAV-BS to best serve on-ground MC users while maintaining a good backhaul connection. System-level simulations are performed to gain insights into the impact of different optimizing parameters on the considered system performance, i.e., the throughput and drop rate of MC users. The simulation data has also been utilized for the RL algorithm design and validation.

\section{Use Case and System Model}

We consider a multi-cell mobile cellular network as illustrated in the right plot of Figure \ref{fig:DeploymentScenario}. The network initially consists of seven macro-BSs. However, due to, for example, a natural disaster, the macro-BS located in the middle of the network map is damaged. Hence, a UAV-BS is temporally set up to provide wireless connectivity to the MC users in the disaster area (a circle area with a 350 m radius in the middle of the deployment map). 
The UAV-BS is modelled as an IAB node. To reduce the complexity and weight of antennas put on the UAV-BS, we assume that the same antennas are used for wireless access and backhaul links. The UAV-BS measures the wireless links to the six functioning macro-BSs, and it dynamically selects one of these macro-BSs that gives the best link quality as its donor node. Then, a wireless backhaul link is established between the UAV-BS and the selected donor-BS. Both normal users and MC users are allowed to access the UAV-BS. A user selects its serving BS (a macro-BS or a UAV-BS) based on the end-to-end wireless path quality.

It is assumed that all macro-BSs and the UAV-BS have three sectors each, and they operate at the same carrier frequency of 3.5 GHz with a time division duplex (TDD) pattern that consists of four time slots, i.e.,  downlink (DL), DL, uplink (UL) and DL. The pattern is repeated with a periodicity of 2 ms. The 100 MHz total system bandwidth is shared between the access and backhaul links. To reduce the complexity and mitigate interference, we further assume that the UAV-BS operates in a half-duplex mode, i.e., it cannot transmit and receive signals simultaneously. The UAV-BS's flying height is assumed to be below 35 meters so that the rural macro propagation model can be reused for UAV-BS in this case.

Users are randomly dropped in the map shown in Figure \ref{fig:DeploymentScenario}. In each time slot, a number of users are activated following a dynamic traffic model with a predefined traffic arriving rate and a predefined average traffic size. The DL and UL traffic of activated users are scheduled based on the access and backhaul link quality, the network scheduling strategy, and the allowed transmission directions at a given time slot at each BS. The throughput of each served user is calculated based on its served traffic size and the time used for delivering the traffic. Note that for a user connected to the UAV-BS, its throughput depends not only on the access link between itself and the UAV-BS but also on the wireless backhaul link between the UAV-BS and the donor-BS. A user will not be served with more traffic than required, and a user can also be dropped/blocked in case of poor link quality or insufficient radio resources. User throughput and drop rate are the key performance indicators considered in our RL algorithm design.

\begin{figure}[t]
  \begin{center}
    \includegraphics[scale=0.72]{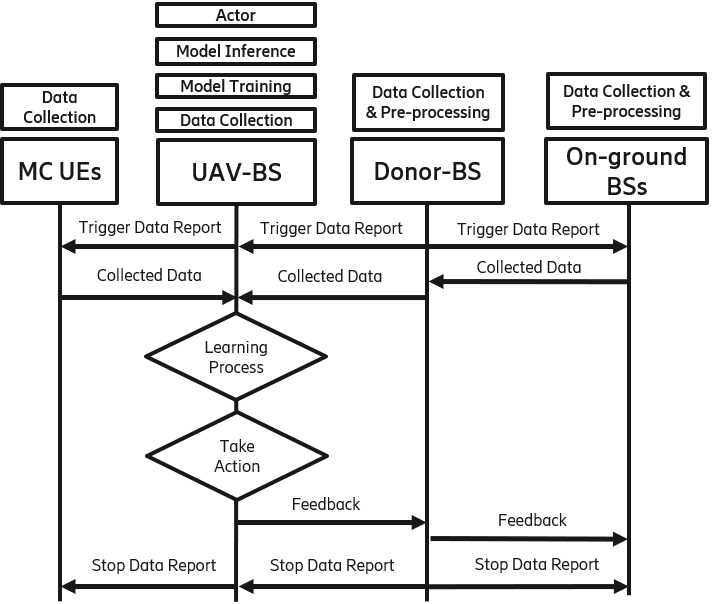}
    \setlength{\abovecaptionskip}{-12pt}
    \caption{Framework and signaling procedure}
    \vspace{-20pt}
    \label{fig:Framework-SignalingProcedure}
  \end{center}
\end{figure}

\begin{figure*}[t]
  \begin{center}

    \includegraphics[scale=0.35]{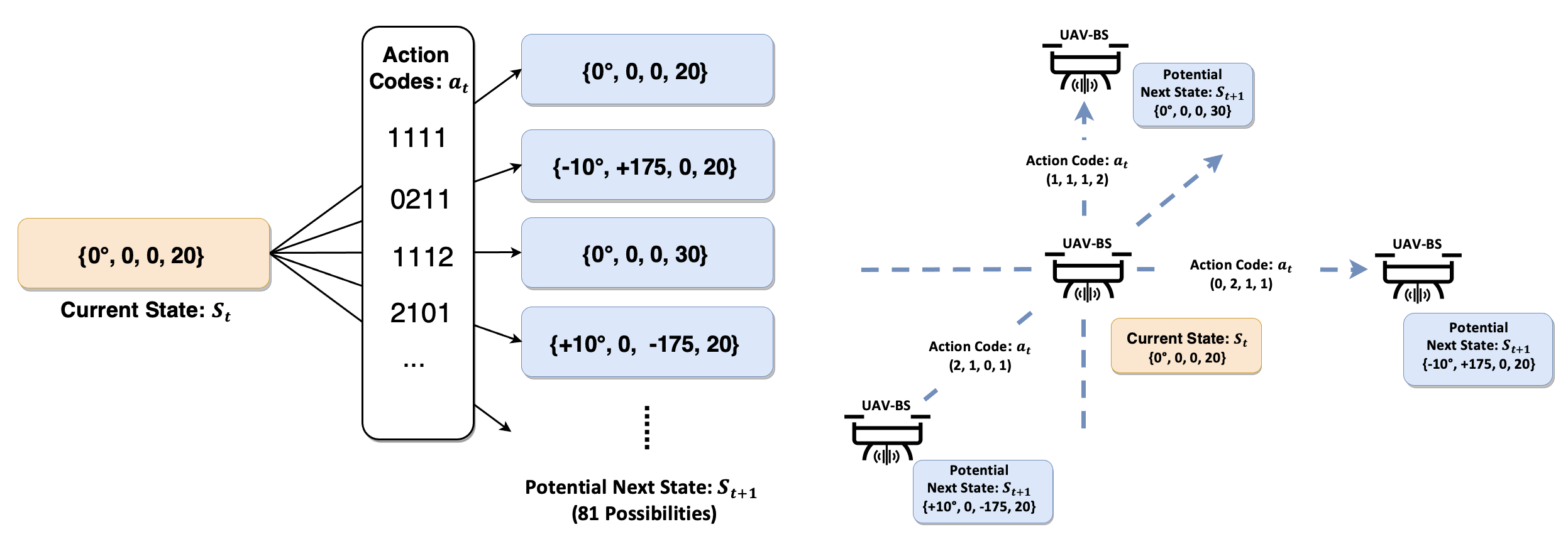}
    \vspace{-10pt}
    \caption{Example of the state transition from current state $\{0\degree, 0, 0, 20\}$}
    \vspace{-25pt}
    \label{fig:action}
    
  \end{center}
\end{figure*}

\section{Framework and Signaling Procedure}
In this section, we propose a signalling procedure to support applying RL to the considered use case. The functional framework studied in 5G NR for radio access network intelligence is used as our reference \cite{3gpp.37.817}.

The blocks above each entity (UE or BS) shown in Figure \ref{fig:Framework-SignalingProcedure} denote different machine learning functionalities, including data collection, model training, model inference and actor. Data collection is a function that is responsible for collecting input data for model training and model inference functions. The model training function performs the training of the learning model while the model inference function provides the learning output. Finally, the actor function receives the output from the model inference module and triggers or performs corresponding actions. Figure \ref{fig:Framework-SignalingProcedure} shows the proposed signalling procedure, which consists of the following key steps:

\begin{enumerate}[1):]

\item Data requests triggered by a donor BS: After a UAV-BS completes its network integration procedure, the donor-BS triggers data collection to assist the UAV-BS in optimizing its configuration and deployment by sending a data request message to the relevant users and BSs. 

\item Data collection at UAV-BS: After receiving the data request, MC users, the donor-BS, and related on-ground BSs will send the requested data to the UAV-BS. The data collection procedure can include: a) data collected by the UAV-BS itself, i.e., from its connected MC users, radio measurements, and onboard sensors. b) data is firstly reported from a set of users and a set of on-ground BSs to the donor-BS and then forwarded to the UAV-BS. 

\item Learning process in the UAV-BS: The UAV-BS processes the collected data. And the model training and inference functions are initiated using the processed data. After training, a set of ML-related parameters are generated corresponding to the trained model. Then, the deployment strategy and configuration are developed based on the output of model inference. 

\item The UAV-BS takes action: The actor function in the UAV-BS performs antenna tilt and location adjustment based on the output of the model inference. 

\item The UAV-BS sends feedback to its donor-BS, who can then forward the feedback or action recommendations to related on-ground BSs. 

\item The donor-BS and/or the related BSs adjust their configuration (e.g., antenna tilt, transmit power, etc.), using the feedback from the UAV-BS as input data. Finally, the donor-BS stops requesting data. 
\end{enumerate}

Steps 2)-6) can repeat till certain criteria are fulfilled. The donor-BS then can stop the learning process by sending a stop data reporting message to its connected users and BSs.

\vspace{-3pt}
\section{Reinforcement Learning Algorithm Design}
\vspace{-3pt}

In this section, we design an RL algorithm to jointly optimize the access and backhaul antenna tilt value and the 3-D location of the UAV-BS in the considered scenario. We use a deep Q-Network as our base algorithm \cite{mnih2015human}. The algorithm is modified and implemented to solve our system optimization problem.



\begin{figure*}[b]
\centering
\subfigure[Antenna Tilt vs Backhaul Link Rate]{
\begin{minipage}[b]{0.3\textwidth}
\vspace{-15pt}
\includegraphics[width=\textwidth]{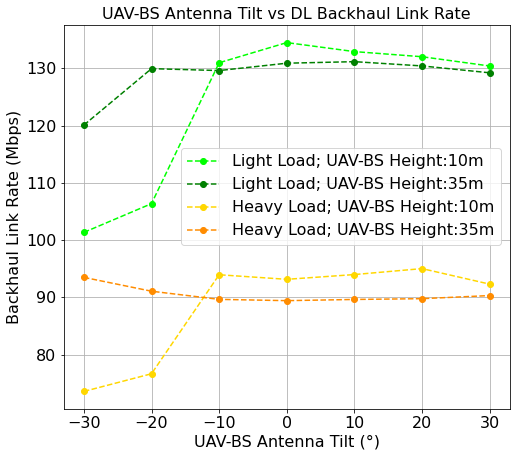} 
\vspace{-15pt}
\label{fig:Tilt:a-BackhaulLinkRate}
\end{minipage}
}
\subfigure[Antenna Tilt vs Throughput for MC Users]{
\begin{minipage}[b]{0.3\textwidth}
\vspace{-15pt}
\includegraphics[width=\textwidth]{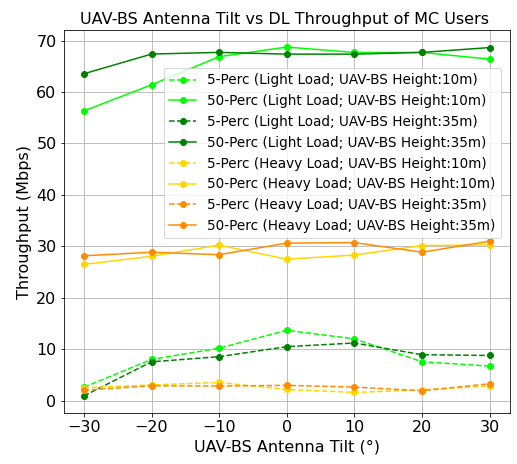}
\vspace{-15pt}
\label{fig:Tilt:b-Throughput}
\end{minipage}
}
\subfigure[Antenna Tilt vs Drop Rate for MC Users]{
\begin{minipage}[b]{0.3\textwidth}
\vspace{-15pt}
\includegraphics[width=\textwidth]{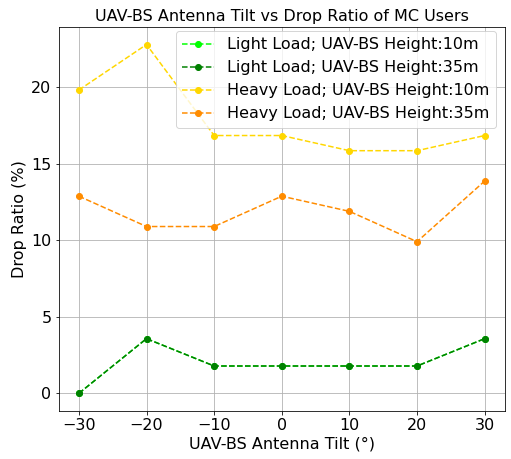}
\vspace{-15pt}
\label{fig:Tilt:c-DropRate}
\end{minipage}
}
\vspace{-5pt}
\caption{UAV-BS antenna tilt's impact on backhaul link rate, MC user throughput and MC user drop rate}
\vspace{-18pt}
\label{fig:Tilt}
\end{figure*}

\vspace{-5pt}
\subsection{Algorithm Environment}

\subsubsection{State Space}

A UAV-BS's state at a given time instance $t$ has four dimensions and it is denoted as $s_t=\{\sigma_t, x_t, y_t, z_t\}$, where $\sigma_t$ represents the electrical tilt value of the access and backhaul antenna, and $\{x_t, y_t, z_t\}$ denotes the 3-D location of the UAV-BS at time $t$. The candidate values for the $x$ and $y$ axis are $[-350,-175, 0, +175, +350]$ meters, which covers the disaster area shown in Figure \ref{fig:DeploymentScenario}. The candidate values of $z$ axis are $[10, 20, 30, 35]$ meters. The candidate antenna tilt values are $[-30, -20, -10, 0, +10, +20, +30]\degree$, where a positive tilt value means applying an electrical down-tilt to the access and backhaul antenna, and a negative tilt value maps to applying an electrical up-title to the antenna.

\subsubsection{Action Space}

For each state dimension, the UAV-BS can select an action out of three candidate options. These three alternative action options are coded by three digits $\{0, 1, 2\}$, where ``0'' denotes that the UAV-BS reduces the status value by one step from its current value; ``1'' represents that the UAV-BS does not need to take any action at this state dimension and it keeps the current value; and ``2'' means that the UAV-BS increases the status value by one step from its current value. For instance, if the UAV-BS is at the space point where the value of the $x$ dimension is equal to 0 meter, then, an action coded by ``0'' for this dimension means that the UAV-BS will select an action to reduce the value of $x$ axis to -175 meters, an action coded by ``1'' implies that the UAV-BS will hold the current value of $x$ axis (0 meter), and an action coded by ``2'' implies that the UAV-BS will increase the value of $x$ axis to 175 meters. The same policy is used for all the dimensions of the state space. Since one state has four dimensions and each state dimension has three action options, the action pool contains in total 81 action candidates that can be programmed to a list of $[0000, 0001, 0002, 0010..., 2222]$. Hence, at a given time $t$, the UAV-BS can select an action $a_t$ from these 81 candidates. Figure \ref{fig:action} demonstrates an example of state transition from a given state $s_t=\{0\degree, 0, 0, 20\}$.

\subsubsection{Monitoring Feature Metrics and Reward Function}

For MC communications, it is more important to serve as many MC users as possible with adequate service quality rather than maximizing the peak rate of a subset of MC users. Hence, for the reward function design of the RL algorithm, we have chosen six key feature metrics to reflect the overall quality of service for MC users, including:
\vspace{-1pt}
\begin{itemize}

\item The drop rates of MC users for 
UL and DL ($\beta_{ul}, \beta_{dl}$), which reflect the percentage of unserved MC users.

\item The 50-percentile throughput values of MC users for both UL and DL ($\alpha_{ul-50\%}  \alpha_{dl-50\%}$), which represent the average performance of the MC users, and

\item The 5-percentile throughput values of MC users for both UL and DL ($\alpha_{ul-5\%}, \alpha_{dl-5\%}$), which represent the ``worst" performance of the MC users.

\end{itemize}
\vspace{-1pt}
To balance these key performance indicators, the reward function is designed as a weighted sum of these six feature values as follows. All features are normalised within the range $[0, 1]$ before model training.
\vspace{-2pt}
\begin{equation}
\NORMAL{
\begin{aligned}
R_s = & \omega_1 \times \frac{(1 - \beta_{dl}) + (1 - \beta_{ul})}{2} + \omega_2 \times \frac{(\alpha_{ul-5\%} + \alpha_{dl-5\%})}{2}\\
& + \omega_3 \times \frac{(\alpha_{ul-50\%} +  \alpha_{dl-50\%})}{2}\\
\end{aligned}
}
\end{equation}
\vspace{-5pt}

In addition, we set $ \omega_1 + \omega_2 + \omega_3 = 1$ to normalise the reward value such that $R_s$ is between $[0, 1]$. The weights assigned to each metric signify the system's relevance. In our scenario, we placed higher weights on user drop rates and 5-percentile MC-user throughput features in order to serve and guarantee acceptable service to all MC users. Thus, the weight values used in our algorithm are $\omega_1 =0.5$, $\omega_2=0.3$ and $\omega_3=0.2$.



A deep Q-network is used as our base algorithm to achieve better self-control decisions for the autonomous UAV-BS. At each training episode, the UAV-BS explores the state space and performs Q-value iterations. An $\epsilon$-greedy exploration is applied when determining the action to take at the next time instance. The probability of exploration is given by parameter $\epsilon$. The exploration probability specifies the likelihood that the agent will execute state exploration and choose actions at random. Otherwise, the agent will perform the action that is believed to yield the highest expected reward. The data for each training step is stored in a replay batch $\mathcal{D}$. Specifically, each row of $\mathcal{D}$ contains the tuple $(s_t, a_t, r_{t}, s_{t+1})$, namely, current state, action, reward and next state. Samples will then be randomly selected and used for Q value model updating.


\section{Performance Evaluation}


In this section, we firstly investigate the impact of the antenna configuration and 3-D location of the UAV-BS on the performance of MC users in terms of throughput and drop rate by using system-level simulations. Then, we evaluate the performance of the proposed RL algorithm using the data generated from the simulator.


\begin{figure}[t]
\centering
\subfigure[UAV-BS Position vs 5 Percentile Throughput for MC Users]{
\begin{minipage}[b]{0.46\textwidth}
\includegraphics[width=\textwidth]{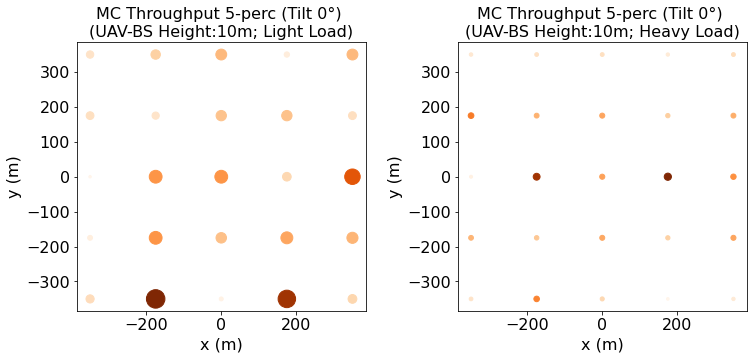} 
\vspace{-15pt}

\label{fig:Position:a-5PercThroughput}
\end{minipage}
}
\subfigure[UAV-BS Position vs 50 Percentile Throughput for MC Users]{
\begin{minipage}[b]{0.46\textwidth}
\includegraphics[width=\textwidth]{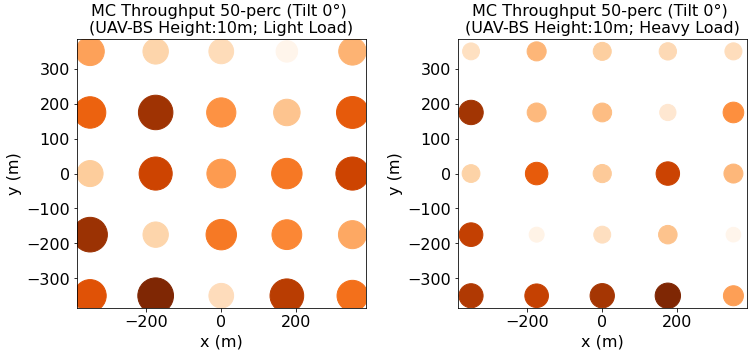}
\vspace{-15pt}
\label{fig:Position:b-50PercThroughput}
\end{minipage}
}
\subfigure[UAV-BS Position vs Drop Rate for MC Users]{
\begin{minipage}[b]{0.46\textwidth}
\includegraphics[width=\textwidth]{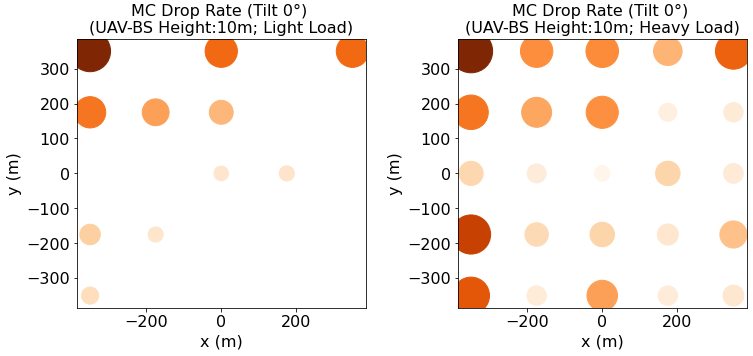}
\vspace{-15pt}
\label{fig:Position:c-DropRate}
\end{minipage}
}
\vspace{-5pt}
\caption{Impact of UAV-BS position on MC user throughput and drop rate}
\vspace{-20pt}
\label{fig:Position}
\end{figure}

\vspace{-5pt}
\subsection{System Performance Analysis}

We consider the system model discussed in section II. It is assumed that all BSs and users have two transmitting and two receiving antennas. The maximum allowed transmit powers for a macro-BS, a UAV-BS and a user are configured as 46, 40, and 23 dBm, respectively. Due to space limitations, in this subsection, we only discuss the DL performance results, considering two different levels of traffic load in the system (light and heavy load cases with different user arriving rates). However, both DL and UL metrics discussed in section III have been used when evaluating the proposed RL algorithm.







To gain insights into the impact of UAV-BS antenna tilt and flying height on the considered performance metrics, we fix the UAV-BS's 2-D position in the center of the deployment map and investigate the system performance in terms of the backhaul link rate, DL throughput and drop rate of MC users for different traffic load levels, as shown in Figure \ref{fig:Tilt}. It can be seen from Figure \ref{fig:Tilt:a-BackhaulLinkRate} that both UAV-BS antenna tilt and height have a significant impact on the backhaul link rate, while UAV-BS height has a bigger impact in the heavy load case. Another observation is that the network has a lower backhaul link rate in case of heavy load. This is because of the increased interference levels, and also the UAV-BS needs to share the resource with the on-ground users connected to its donor-BS (in-band IAB operation). Hence, increasing traffic load means that more users will share the radio resource with the UAV-BS for its backhaul traffic delivery. 

From Figure \ref{fig:Tilt:b-Throughput}, we see that the impact of tilt values on the throughput performance in the light load case is more visible, while for the high load case, the impact of tilt values is negligible. Meanwhile, UAV-BS flying height has no significant impact on the throughput in the considered UAV-BS 2-D location. 
Figure \ref{fig:Tilt:c-DropRate} shows that both antenna tilt and UAV-BS height significantly impact drop rate performance. If the system load is light, the curve for the UAV-BS height at 10 m overlaps with the curve for the case of 35 m. This implies that in the current context, UAV-BS height has no discernible effect on the drop rate of MC users, while UAV-BS antenna tilt plays a more prominent role. For the heavy load situation, the results imply that flying the UAV-BS at a higher altitude can reduce the drop rate.

\begin{figure}[t]
  \begin{center}

    \includegraphics[scale=0.42]{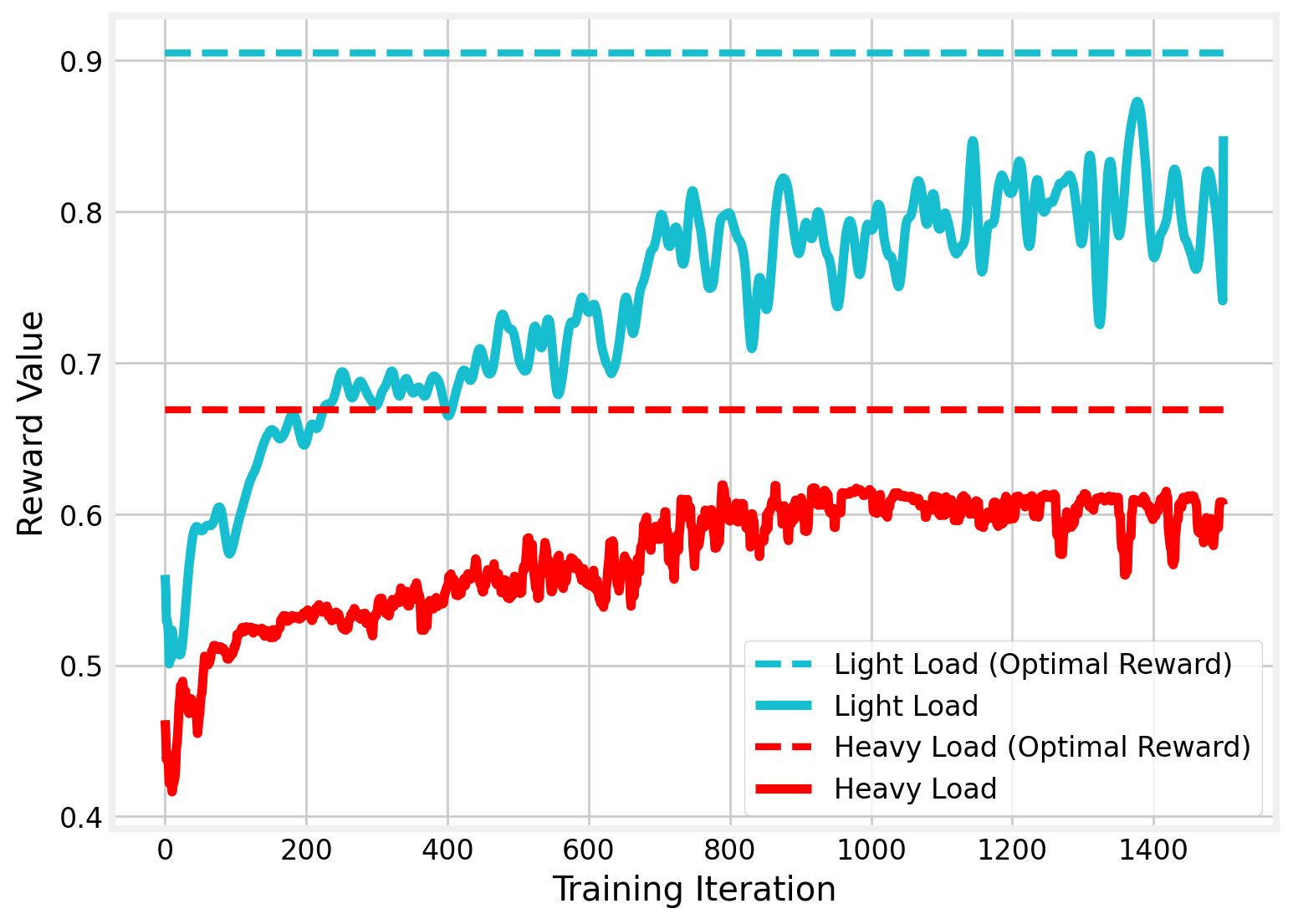}
    \vspace{-20pt}
    \caption{Reward value with number of training iterations in two different network traffic load scenarios}
    \vspace{-23pt}
    \label{fig:reward}
  \end{center}
\end{figure}

\begin{table*}[t]
\centering

 \caption{Comparison of the average value of the six system performance metrics during UAV-BS deployment in two load scenarios}

\label{tab:re3}
 \begin{tabular}{c C{0.6in} C{0.6in} C{0.6in} C{0.6in} C{0.6in} C{0.6in} C{0.6in}} 

\toprule
 & DL 50\textperthousand \quad Throughput  & DL 5\textperthousand \quad Throughput & DL Drop Percentage & UL 50\textperthousand \quad Throughput  & UL 5\textperthousand \quad Throughput & UL Drop Percentage & Reward Value \\
\midrule
Optimal State (Light Load) & 106.11 & 26.03 & 0\%  & 7.73 & 2.18 & 2.4\% & 0.905\\ 
\midrule
\makecell{Reinforcement Learning\\(Light Load)} & 105.47 & 25.56 & 0.1\%  & 7.7 & 2.15 & 2.6\% & 0.865\\ 
\midrule
\midrule
Optimal State (Heavy Load) & 35.5 & 3.51 & 8.9\% & 4.68 & 0.15 & 2\% & 0.669\\
\midrule
\makecell{Reinforcement Learning\\(Heavy Load)} & 58.4 & 1.55 & 12\% & 5.08 & 0.1 & 3\% & 0.617\\
\bottomrule

\end{tabular}
\vspace{-18pt}
\end{table*}

Figure \ref{fig:Position} shows the impact of UAV-BS's 2-D position (x-axis and y-axis) on the considered performance metrics when the UAV-BS antenna tilt is fixed to 0° and its flying height is fixed to 10 m. The size of the circles shown in Figures \ref{fig:Position:a-5PercThroughput}, \ref{fig:Position:b-50PercThroughput} and \ref{fig:Position:c-DropRate} represents the exact value of the 5-percentile throughput, 50-percentile throughput and drop rate of MC users, respectively. The larger the size is, the higher value is for a considered performance metric. To make it easier to identify the 2-D location that gives the highest value of a considered performance metric, in each subplot of Figure \ref{fig:Position}, we use different colours of a circle to represent a relative value of the considered performance metric. The darker the colour is, the higher value of the performance metric is. 

It can be seen from Figure \ref{fig:Position} that for a given performance metric, e.g., 5-percentile MC user throughput, the optimal UAV-BS 2-D location changes when the network traffic load level changes. We can also observe that the optimal UAV-BS position is close to the edge instead of the center of the MC area. This is because the UAV-BS can keep a good backhaul connection with its donor-BS at the edge of the MC area. In addition, we see that the optimal UAV-BS position is different when considering different performance metrics, e.g., maximizing the 5-percentile MC user throughput, maximizing the 50-percentile MC user throughput, or minimizing the MC user drop rate. Therefore, the weights selected for different performance metrics in the reward function will impact the optimal location of the UAV-BS.

\vspace{-2pt}
\subsection{Reinforcement Learning Performance Evaluation}
\vspace{-2pt}

In this section, we present the results obtained through our RL algorithm proposed in Section IV. We compare the result of our algorithm with the global optimal state, which is obtained by using grid search through the whole data set. We show the benefits of using the proposed algorithm and its ability to adapt to the changing wireless environment after model training. During the training, the learning rate is set to $5 \times 10^{-5}$, the initial starting exploration probability $\epsilon$ is set to 1 while the exploration decay is 0.995 and the number of training iterations equals 1500.

We first analyze the convergence of the algorithm during the model training. Figure \ref{fig:reward} illustrates the reward value as a function of the number of training iterations. We can observe that, for both the light load and heavy load cases, the proposed algorithm can learn from the history and eventually approach the optimal state that gives the largest reward value in both load scenarios. The same conclusion can also be made by comparing the reward value column of Table \ref{tab:re3}.

Table \ref{tab:re3} also shows the performance of the considered six feature metrics during the UAV deployment. Each deployment contains 100 steps for a UAV-BS to make decisions and take action. We can observe that based on past experience and a well-trained learning model, the proposed algorithm can quickly configure and navigate the UAV-BS to optimize the considered performance metrics. Only a limited number of steps are needed for the UAV-BS to reach a stable state. As shown in Table \ref{tab:re3}, the reward value (a weighted sum of the six considered performance metrics) achieved by the proposed RL algorithm is only about 4\% to 5\% less than that provided by the global optimal solution for the light load and high load scenarios, respectively. Since the reward value summarizes the overall system performance metrics during the UAV-BS deployment, our results demonstrate the strength of the algorithm and the ability to provide fast connectivity to MC users in different traffic load scenarios.

\vspace{-2pt}
\section{Conclusions and Future Work}
\vspace{-2pt}

In this paper, we developed an RL algorithm to autonomously configure and navigate a UAV-BS to provide temporary coverage for MC users. The UAV-BS is connected to an on-ground donor BS and integrated into an existing mobile network using the 5G IAB technology. A functional framework and signalling procedure are proposed to support data collection, model training and decision making for the considered use case. An action encoding strategy is introduced to represent UAV-BS decisions with multiple state dimensions, including the 3-D space location and the access and backhaul antenna electrical tilt. Our results demonstrate the benefits and efficiency of our proposed algorithm in different traffic load scenarios. The algorithm can help a UAV-BS quickly find the optimal 3-D location and its antenna configuration to provide a stable connection to MC users.

In the future work, we will further investigate various hyper-parameter and reward function combinations based on different service requirements. We will also investigate various frameworks and signaling procedures to support the application of centralized or distributed machine learning for the considered use case.

\vspace{-2pt}

\bibliographystyle{IEEEtran}
\bibliography{cite}

\end{document}